\documentclass[11pt]{article}
\usepackage[left=1in,right=1in,top=1.2in,bottom=1.2in,
            footskip=.25in]{geometry}
\usepackage{amsmath, amssymb,amstext} 
\usepackage{url}
\usepackage{amsmath}
\usepackage[table]{xcolor}
\usepackage{authblk}
\usepackage{amssymb}
\usepackage{ragged2e}
\usepackage{hyperref}
\usepackage{graphics}
\usepackage{mathtools}
\usepackage{caption}
\usepackage{subcaption}
\usepackage{algpseudocode}
\usepackage{algorithm}

\usepackage{afterpage}
\usepackage{makecell}
\usepackage{natbib}
\usepackage{array}
\usepackage{collcell}
\usepackage{tcolorbox}
\usepackage{amsmath,amssymb}
\usepackage{graphicx}
\usepackage{subcaption}
\usepackage{enumitem}
\usepackage{booktabs}
\usepackage{multirow}
\usepackage{algorithm}
\usepackage{algpseudocode}
\usepackage{natbib}
\usepackage{amsthm}

\title{Learning PDE Time-Stepping with Neural Cellular Automata}
\author[1,a]{Esha Saha}
\author[1,a]{Hao Wang}

\affil[1]{Interdisciplinary Lab for Mathematical Ecology and Epidemiology (ILMEE) \& The Department of Mathematical and Statistical Sciences, University of Alberta, Edmonton (AB), T6G 2J5, Canada}
\affil[a]{Corresponding authors. E-mail: esaha1@ualberta.ca, hao8@ualberta.ca}
\date{}
\begin{document}
\maketitle

\begin{abstract}
Classical numerical solvers for partial differential equations (PDEs) are computationally expensive to solve repeatedly across varying initial conditions, motivating the need for learned surrogates. In this paper, we propose a trainable Neural Cellular Automata (NCA) based surrogate model for learning long time PDE dynamics. 
Rather than mapping an entire initial field to a full trajectory in one shot, our proposed model learns a small, local, homogeneous update rule that is applied identically and repeatedly at every grid cell, mirroring the locality of differential operators. We benchmark this framework against three baselines: PDE - Net, a modified physics-informed neural network (PINN), and a Fourier Neural Operator (FNO), on five canonical PDEs (heat, advection, Burgers, Allen - Cahn, and Fisher - KPP), evaluated at temporal domain two times beyond the training temporal domain. The proposed model achieves the lowest long-horizon relative errors on the majority of the experiments. 
\end{abstract}
\section{Introduction}

Partial differential equations (PDEs) govern the dynamics of a vast range of physical phenomena, from heat conduction and fluid flow to reaction - diffusion processes in biology and materials science. 
Classical numerical methods such as finite difference, finite element, and spectral solvers remain the backbone for simulating such systems, offering strong convergence guarantees and interpretability \cite{ames2014numerical}.
However, these solvers can be computationally expensive, especially in settings where the same equation must be solved repeatedly for varying initial conditions, boundary conditions, or coefficients. This has motivated a growing body of work on \emph{learned surrogates}: neural networks trained to approximate the solution operator of a PDE directly from data, reducing the cost of simulation across many queries \cite{blechschmidt2021three,luo2025physics,lagaris1998artificial}.

Inspired by the works of \cite{richardson2024learning} on learning spatiotemporal dynamics using cellular automata, in this work, we explore \emph{Neural Cellular Automata} (NCA) as a surrogate for PDE time-stepping. Instead of mapping an entire initial field to an entire trajectory in one shot, the NCA learns a \emph{local}, \emph{homogeneous} update rule using a small neural network applied identically and repeatedly at every grid cell using only its immediate neighborhood \cite{mordvintsev2020growing,richardson2024learning}. 
This mirrors the locality of differential operators and trains a model that is, by construction, translation-invariant.
Repeated application of the learned rule over many steps produces a full spatiotemporal rollout, which is trained to match trajectories generated by a classical numerical solver.

We test our framework on five canonical PDEs, describe the finite-difference discretization used to generate training trajectories and the initial-condition families used to test the learned dynamics, and benchmark the trained NCA against three baseline learned solvers - PDE - Net, a PINN, and FNO on learning solutions for a temporal domain up to $2$ times beyond the training temporal domain
(Section~\ref{sec:results}).

\section{Related Work}
\label{sec:related}

Classical discretization schemes (finite difference, finite element, and spectral methods) remain the basic standard for PDE simulation but have a computational cost that scales with grid resolution and time horizon.
Physics-informed neural networks (PINNs) \citep{raissi2019physics} instead parameterize the solution field directly with a neural network and penalize residuals of the governing equation, but typically require retraining for each new initial or boundary condition and can struggle with stiff or multi-scale dynamics. 
PINNs have been widely used to learn and solve PDEs as well as ODEs widely across the literature \citep{farea2024understanding, de2024physics,matthews2024pinnde,arora2024invariant,toscano2025pinns}. 
Applications of PINNs include fluid dynamics \citep{cai2021physics}, air quality modeling \cite{saha2025dispersion, zhang2026eulerian,li2023physical,kim2025dynamic}, etc. 
Some recent works have also developed sparse PINNs for data-scarce applications \cite{chen2021physics,saha2025learning}.
\paragraph{Neural operators and learned simulators.}
In order to learn models that are independent of training for specific initial/boundary conditions, a separate line of work learns a mapping between function spaces rather than
a single solution instance, often known as operator learning.
The Fourier Neural Operator (FNO) \cite{li2020fourier} parameterizes the solution operator in the spectral domain and generalizes across initial conditions after training, while DeepONet \citep{lu2021deeponet} uses a branch - trunk architecture to approximate general nonlinear operators. 
Closely related to the cellular-automaton framework, graph-network-based simulators \citep{eliasof2021pde,li2020multipole} represent the domain as a mesh or particle graph and learn local message-passing update rules, and have been
extended to learned turbulence and weather models
\cite{stachenfeld2021learned}. 
PDE - Net \cite{long2018pde,long2019pde} learns a bank of \emph{trainable} convolutional filters jointly with a pointwise nonlinearity to approximate the local time-stepping operator directly from data.

\paragraph{Neural Cellular Automata.}
NCA was introduced by \cite{mordvintsev2020growing,hartl2025neural} as a differentiable, locally-updating model capable of growing and regenerating complex target morphologies from a single seed cell, trained end-to-end through a stochastic, asynchronous update rule reminiscent of biological morphogenesis. 
Follow-up work extended NCA to texture synthesis \citep{niklasson2021selforganising} and self-classifying and self-organizing behaviors, demonstrating that a shared, local update function can encode
rich, globally-coherent spatiotemporal behavior. 
Many extensions of NCA have been studied such as its variational formulation \cite{palm2022variational}, attention based NCA \cite{tesfaldet2022attention}, etc.
Richardson et al.\ \cite{richardson2024learning} train NCAs to reproduce spatiotemporal patterns, including Gray--Scott PDE trajectories and image morphing, using local residual updates and a fixed differential-operator perception.
We do not claim a more powerful NCA than \cite{richardson2024learning}, nor a new morphogenesis objective.
Our contribution is to recast that local-update construction as a \emph{learned numerical PDE time-stepper}: a gated finite-difference perception dictionary, a synchronous residual update without stochastic firing, and a long-horizon comparison against PDE - Net, PINN, and FNO under a shared finite-difference solver on five canonical PDEs.
Table~\ref{tab:related} makes this positioning explicit.
We do not evaluate generalization across PDE coefficients, grid resolution, or boundary conditions, and we treat learned gates as stencil preferences rather than recovered governing equations.

\begin{table}[ht]
\centering
\caption{Where this work sits relative to prior NCA and learned PDE solvers.}
\label{tab:related}
\small
\resizebox{\textwidth}{!}{%
\begin{tabular}{@{}lcccc@{}}
\toprule
 & Prior NCA$^{\dagger}$ & PDE - Net & FNO & This work \\
\midrule
Task
  & patterns / Gray--Scott
  & PDE stepper
  & PDE operator
  & PDE stepper \\
Local update
  & $\checkmark$ & $\checkmark$ & $\times$ & $\checkmark$ \\
Stochastic firing mask
  & $\checkmark$ & $\times$ & $\times$ & $\times$ \\
Perception
  & fixed stencils
  & learned filters
  & Fourier modes
  & fixed stencils + gates \\
Long-horizon
  & $\times$ & $\times$ & $\times$ & $\checkmark$ \\
\bottomrule
\end{tabular}%
}
\begin{minipage}{\textwidth}
\smallskip\footnotesize
$^{\dagger}$Mordvintsev et al.\ \cite{mordvintsev2020growing} and Richardson et al.\ \cite{richardson2024learning}.
\end{minipage}
\end{table}
\section{Methodology}
\label{sec:methodology}

\subsection{Neural Cellular Automata}
\label{sec:nca-arch}
\textbf{Notations.}
Let $x\in\mathbb{R}^2$ denote the spatial coordinates and
$t\in[0,T]$ denote the physical time. We denote the physical solution
of a PDE by $u(x,t)$, governed by
\[
\frac{\partial u}{\partial t}=F(u,x,t),
\]
where $\partial u/\partial t$ is the temporal derivative of the
physical field.

Following the Neural Cellular Automata formulation of
\cite{richardson2024learning}, we represent the evolving PDE field using
a multi-channel NCA state rather than directly evolving the physical
field alone. At rollout step $k$, the NCA state is
\[
z^{(k)}\in\mathbb{R}^{B\times C\times H\times W},
\]
where $B$, $C$, $H$, and $W$ denote the batch size, number of channels,
height, and width, respectively. The first channel contains the physical
PDE variable,
\[
u^{(k)} = z^{(k)}_{:,0,:,:},
\]
while the remaining $C-1$ channels are latent states that allow the NCA
to retain additional information, such as implicit derivatives or
temporal context.

\begin{figure}[htbp]
\centering
\includegraphics[width=0.75\textwidth]{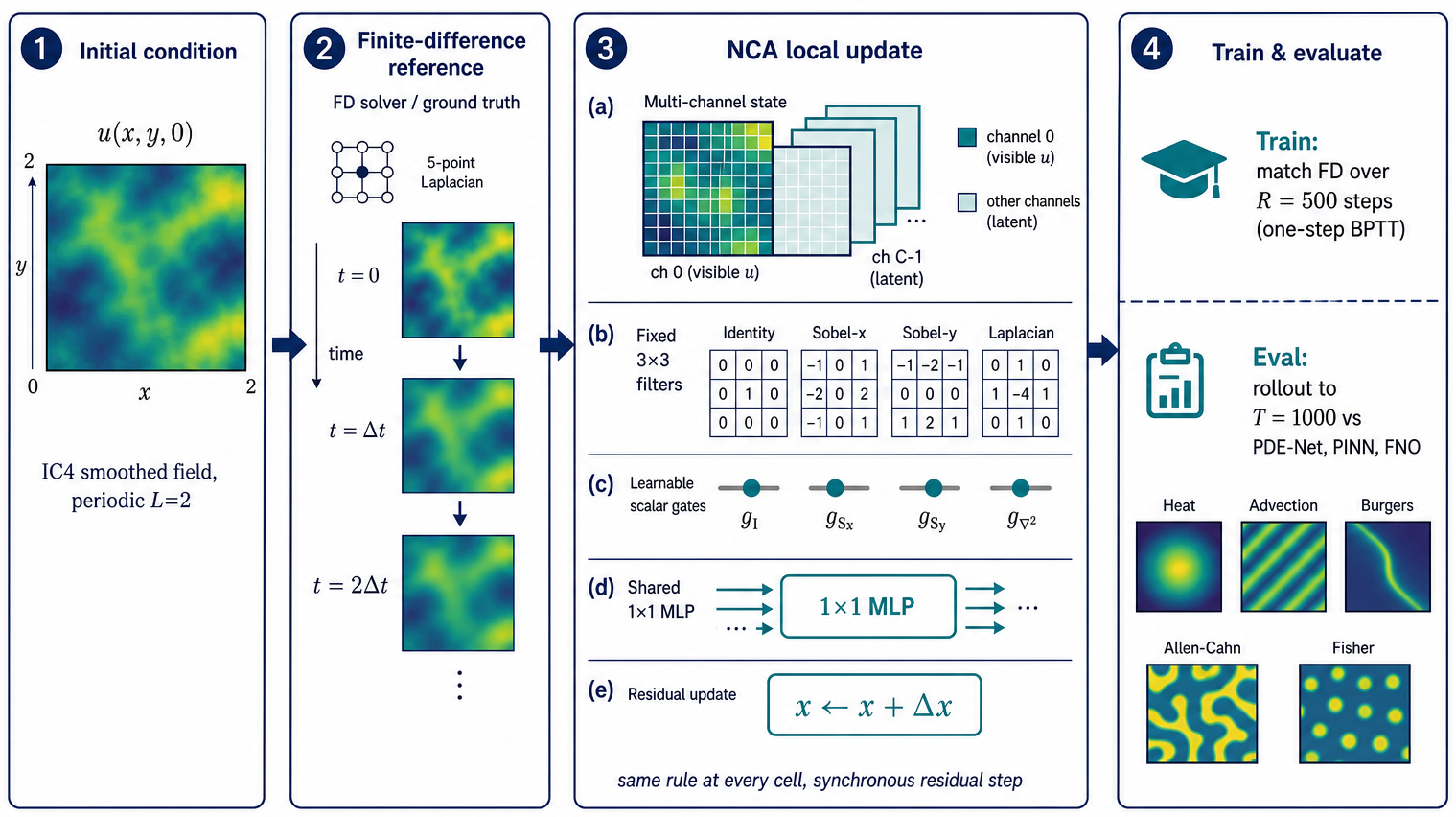}
\caption{One NCA forward pass: fixed-filter perception with learnable
scalar gates, a shared pointwise MLP producing a proposed update, and a
synchronous residual update. Repeated application rolls the state forward
in time and is trained against ground-truth finite-difference trajectories.}
\label{fig:architecture}
\end{figure}

At each rollout step, the NCA first computes a local perception of the
current state,
\[
p^{(k)}=\mathcal{P}\!\left(z^{(k)}\right),
\]
where $\mathcal{P}$ denotes the gated fixed-stencil perception operation
defined below.
The perceived features are then passed through the learned update
network to obtain a $C$-channel state update,
\[
\Delta z^{(k)}=\mathcal{N}_{\theta}\!\left(p^{(k)}\right),
\qquad
\Delta z^{(k)}\in\mathbb{R}^{B\times C\times H\times W}.
\]
The complete NCA state is updated by a synchronous residual step,
\[
z^{(k+1)}
=
z^{(k)}
+
\Delta z^{(k)}.
\]
Unlike classical morphogenesis NCAs, we do \emph{not} apply a stochastic
cell-firing mask: every cell updates at every step.
The physical PDE field after each rollout step is obtained from channel~0,
\[
u^{(k+1)}=z^{(k+1)}_{:,0,:,:}.
\]
Thus, the NCA learns a local transition function over the full
multi-channel state, while only the first channel is interpreted as the
physical PDE solution. The rollout index $k$ denotes the discrete NCA
update step and should not be confused with the continuous physical
time variable $t$ used in the PDE formulation.
Figure~\ref{fig:architecture} illustrates one forward pass, which consists
of three stages:

\begin{itemize}[leftmargin=*]
\item \textbf{Perception.}
A depthwise convolution with four fixed, non-learned $3{\times}3$ filters
is applied independently to every channel (with circular padding),
producing a perception tensor of shape $(B,4C,H,W)$.
The stencil coefficients themselves are not learned; instead, each of the
$4C$ perception channels is multiplied by a trainable scalar gate that
controls its contribution to the update network.
Specifically, we use the identity, Sobel-$x$, Sobel-$y$, and discrete
Laplacian kernels
\[
K_I =
\begin{bmatrix}
0&0&0\\
0&1&0\\
0&0&0
\end{bmatrix},
\qquad
K_x =
\frac{1}{8}
\begin{bmatrix}
1&0&-1\\
2&0&-2\\
1&0&-1
\end{bmatrix},
\qquad
K_y = K_x^{\top},
\qquad
K_{\Delta} =
\begin{bmatrix}
0&1&0\\
1&-4&1\\
0&1&0
\end{bmatrix}.
\]
Writing $g\in\mathbb{R}^{4C}$ for the learnable gates and indexing channels
by $c=0,\ldots,C-1$, the perception map is
\[
\mathcal{P}(z)
=
\operatorname{Concat}_{c,j}
\bigl(
g_{c,j}\,(K_j*z_c)
\bigr),
\qquad
j\in\{I,x,y,\Delta\},
\]
where $*$ denotes spatial convolution.
An $\ell_1$ penalty $\lambda\|g\|_1$ with $\lambda{=}10^{-4}$ is applied
to the gates to encourage a sparse preference among the fixed local stencils (Section~\ref{sec:interpretability}).

\item \textbf{Neural update.}
A pointwise MLP implemented by $1{\times}1$ convolutions maps the gated
perception to a $C$-channel update:
\[
\Delta z
=
W_1\,\operatorname{ReLU}\bigl(W_0\,\mathcal{P}(z)\bigr).
\]
Because the MLP is applied identically at every spatial location, the
resulting rule is translation-invariant.
The output layer $W_1$ is zero-initialized so that training begins as an
identity map, promoting stable early training.

\item \textbf{Residual update.}
The update is added back to the current state,
$z^{(k+1)}=z^{(k)}+\Delta z^{(k)}$,
with no stochastic masking and no learned step-size scaling in the main
multi-PDE experiments.
\end{itemize}

\paragraph{Training}
We distinguish three lengths:
(i)~$10{,}000$ Adam outer steps;
(ii)~a training rollout of $R{=}500$ sequential NCA updates per Adam step;
and (iii)~backpropagation through time (BPTT) truncation length $K{=}1$.
At each Adam step we sample a fresh mini-batch of initial conditions
($B{=}4$).
The NCA state $z^{(k)}$ and the finite-difference scheme $u^{(k)}$ are
advanced in lockstep for $R$ steps.
The next NCA input is the model's own state (no teacher forcing).
After each step we detach $z^{(k)}$ before the next forward pass, so
gradients do not flow across time: each MSE term backpropagates through
only one NCA application (one-step truncated BPTT).
The training loss is the mean MSE between channel~$0$ and the finite difference solver at
every rollout step, plus the gate sparsity penalty $\lambda\|g\|_1$.
Evaluation uses the same closed-loop composition with no detach, out to
horizon $T{=}1000$.
Unlike morphogenesis NCAs, we do not use a stochastic firing mask.

\subsubsection*{Governing PDEs}
\label{sec:pdes}

We consider five canonical PDEs on a two-dimensional grid, spanning diffusive, advective, reactive, and bistable dynamics:
\begin{align}
\text{Heat:} \quad & u_t = \nabla^2 u \\
\text{Burgers:} \quad & u_t + u u_x = \nu u_{xx} \\
\text{Fisher - KPP:} \quad & u_t = \nabla^2 u + r u(1-u)\\
\text{Allen - Cahn:} \quad & u_t = \epsilon \nabla^2 u + u - u^3 \\
\text{Advection:} \quad & u_t + c u_x = 0.
\end{align}
These PDEs are considered with periodic boundary conditions along with the initial condition:
\[
    \tilde{u}_0(x,y)
    =
    (K * \xi)(x,y),
    \]
    where $\xi(x,y)\sim\mathcal{N}(0,1),$
    and \(K\) denotes a \(5\times5\) averaging kernel.
    The normalized field is given by
    \[
    u_0(x,y)
    =
    \frac{
    \tilde{u}_0(x,y)
    }{
    \operatorname{Std}(\tilde{u}_0)+10^{-6}
    }.
    \]
The hidden channels are initialized as $h_k(x,y)\sim0.01\,\mathcal{N}(0,1),\qquad k=1,\dots,C-1.$
Algorithm~\ref{alg:nca-train} summarizes one NCA step and the
one-step-BPTT training / closed-loop evaluation procedures.

\begin{algorithm}[ht!]
\caption{NCA PDE time-stepping: forward step, training, and evaluation}
\label{alg:nca-train}
\begin{algorithmic}[1]
\Function{NCAStep}{$z$}
  \State $p \gets \mathcal{P}(z)$
    \Comment{gated fixed stencils: $I$, Sobel-$x/y$, $\Delta$}
  \State $\Delta z \gets W_1\,\mathrm{ReLU}(W_0\,p)$
    \Comment{pointwise MLP; $W_1$ zero-init}
  \State \Return $z + \Delta z$
    \Comment{synchronous residual; no firing mask}
\EndFunction
\Statex
\Procedure{Train}{$\Phi_{\mathrm{FD}}$, $R{=}500$, $N_{\mathrm{opt}}{=}10^{4}$, $B{=}4$, $\lambda{=}10^{-4}$}
  \For{$s = 1,\ldots,N_{\mathrm{opt}}$}
    \State Sample batch of ICs $\{u_0^{(b)}\}_{b=1}^{B}$ from IC4
    \State Initialize $z^{(0)}$ with channel~$0$ $=$ $u_0$ and hidden channels $\sim 0.01\,\mathcal{N}(0,1)$
    \State $u^{(0)} \gets u_0$;\; $L \gets 0$
    \For{$k = 0,\ldots,R-1$}
      \State $z^{(k+1)} \gets \textsc{NCAStep}(z^{(k)})$
      \State $u^{(k+1)} \gets \Phi_{\mathrm{FD}}(u^{(k)})$
        \Comment{FD solver one step}
      \State $L \gets L + \mathrm{MSE}\!\bigl(z^{(k+1)}_{:,0,:,:},\,u^{(k+1)}\bigr)$
      \State $z^{(k+1)} \gets \mathrm{Detach}(z^{(k+1)})$
        \Comment{BPTT truncation $K{=}1$}
    \EndFor
    \State $L \gets L/R + \lambda\|g\|_1$
    \State Adam update on $(W_0,W_1,g)$ using $\nabla L$
  \EndFor
\EndProcedure
\Statex
\Procedure{Evaluate}{$u_0$, $T_{\max}{=}1000$}
  \State Initialize $z^{(0)}$ from $u_0$ as above;\; $u^{(0)} \gets u_0$
  \For{$k = 0,\ldots,T_{\max}-1$}
    \State $z^{(k+1)} \gets \textsc{NCAStep}(z^{(k)})$
      \Comment{closed-loop; no detach}
    \State $u^{(k+1)} \gets \Phi_{\mathrm{FD}}(u^{(k)})$
    \State $\rho^{(k+1)} \gets
      \|z^{(k+1)}_{:,0,:,:}-u^{(k+1)}\|_2
      \big/
      \|u^{(k+1)}\|_2$
  \EndFor
  \State \Return $\{\rho^{(T)}\}$ for $T\in\{100,200,600,800,1000\}$
\EndProcedure
\end{algorithmic}
\end{algorithm}
\subsubsection*{Experimental Setup}
\label{sec:setup}

The equations are solved on a $64\times64$ grid with $x\in[0,2]\times [0,2]$ (i.e., $\Delta x = 1/32$) with appropriate initial conditions (smooth random field obtained by Gaussian noise). 
We train the model using the spatial snapshots obtained above with a training temporal horizon of $T=500$ (the final time $T$ depends on the choice of $\Delta t$ which varies based on the CFL condition for the PDE).
For training, each Adam step samples a fresh batch of ICs and advances the finite difference update and model.
Post training, we evaluate the learned model based on the following: (i) accuracy of full-field learned solutions of the PDEs from held-out ICs from the training IC family; (ii) temporal extrapolation from training rollout of $T=500$ to horizon $T=1000$ (note that it does not evaluate on unseen PDE coefficients, alternative boundary conditions, resolution transfer, or qualitatively different IC families as a primary claim).
For the PINN baseline, to make the architecture similar to our model, we train a convolutional neural network (CNN) with a combined data-fit and finite-difference physics-residual loss.
This modification was adopted due to the original PINN formulation performing significantly worse than the CNN architecture.
The model is evaluated based on the relative $\ell_2$ error defined by $\|\hat u - u\|_2 / \|u\|_2$, where $\hat u$ and $u$ denote the learned and true solutions respectively, i.e., for a particular temporal horizon we compute the errors between the true and learned solution for each time step.
We report the mean $\pm$ std relative $\ell_2$ errors over 200 different ICs (from the same family) for different time horizons $T\in\{100,200,600,800,1000\}$. For ablation studies and noise robustness experiments, 50 held-out from the family of ICs are used for evaluation. 
We do not claim coefficient, boundary-condition, resolution, or IC-family
transfer as a primary result.
PDE parameters and time steps are listed in Table~\ref{tab:oracle};
shared training settings are listed in Table~\ref{tab:train}.
For full reproducibility, the codes (including baseline implementations will be made available at https://github.com/esha-saha.

\begin{table}[ht]
\centering
\caption{Finite-difference solver for the main multi-PDE benchmark
($N{=}64$, $L{=}2$, periodic boundaries).}
\label{tab:oracle}
\small
\begin{tabular}{@{}lll@{}}
\toprule
PDE & Parameters & Discrete $\Delta t$ \\
\midrule
Heat
  & $u_t=\Delta u$
  & $0.2\,\Delta x^2$ \\
Burgers
  & $\nu{=}0.1$; upwind flux
  & $\min\!\big(0.4\Delta x/(|u|_{\max}{+}\varepsilon),\,0.2\Delta x^2/\nu\big)$ \\
Fisher - KPP
  & $r{=}1$
  & $\min(0.2\Delta x^2,\,0.5)$ \\
  Allen--Cahn
  & $\varepsilon{=}10^{-3}$; reaction $u-u^3$
  & $\min(0.2\Delta x^2/\varepsilon,\,10^{-3})$ \\
  Advection
  & $c{=}1$; upwind (baselines)
  & $0.5\,\Delta x$ (CFL $=0.5$) \\
\bottomrule
\end{tabular}
\end{table}

\begin{table}[ht]
\centering
\caption{Training and evaluation settings for the main benchmark
(Table~\ref{tab:full}). Ablations use $50$ eval seeds; noisy runs use
$50$ eval seeds for $\sigma{>}0$.}
\label{tab:train}
\small
\resizebox{\textwidth}{!}{%
\begin{tabular}{@{}ll@{}}
\toprule
Setting & Value \\
\midrule
Grid / domain
  & $N{=}64$, $L{=}2$, $\Delta x{=}1/32$, periodic BC \\
Initial conditions
  & IC4 smooth Gaussian RF (online sampling) \\
Optimizer
  & Adam \\
Learning rate
  & NCA / PDE - Net: $10^{-4}$; FNO: $5{\times}10^{-4}$; PINN: $10^{-3}$ \\
Eval horizons
  & $T\in\{100,200,600,800,1000\}$ (closed-loop; $200$ seeds) \\
BPTT truncation $K$
  & $K{=}1$ (detach every step; one-step BPTT) \\
NCA capacity
  & $C{=}8$ channels, hidden width $128$ ($\sim$5.3k params) \\
PDE - Net
  & $8$ learnable $5{\times}5$ filters, hidden $64$; same $R{=}500$ loop \\
FNO
  & Fourier stepper ($\sim$2.36M params); same $R{=}500$ detach loop \\
PINN
  & IC-conditioned grid CNN ($\sim$208k params); data + IC + physics residual \\
\bottomrule
\end{tabular}%
}
\end{table}
\section{Results}
\label{sec:results}
\begin{table}[ht]
\centering
\caption{Mean relative $\ell_2$ error.
Bold marks the best model per PDE.}
\label{tab:full}
\small
\resizebox{\textwidth}{!}{%
\begin{tabular}{@{}llccccc@{}}
\toprule
PDE & Model & $T{=}100$ & $T{=}200$ & $T{=}600$ & $T{=}800$ & $T{=}1000$ \\
\midrule
\multirow{4}{*}{Heat}
  & NCA      & $\mathbf{0.011{\pm}0.002}$ & $\mathbf{0.035{\pm}0.011}$ & $\mathbf{0.219{\pm}0.132}$ & $\mathbf{0.369{\pm}0.266}$ & $\mathbf{0.561{\pm}0.486}$ \\
  & PDE - Net  & $0.067{\pm}0.012$ & $0.080{\pm}0.021$ & $0.292{\pm}0.134$ & $0.479{\pm}0.274$ & $0.709{\pm}0.493$ \\
  & PINN     & $0.550{\pm}0.082$ & $0.682{\pm}0.109$ & $1.085{\pm}0.289$ & $1.226{\pm}0.467$ & $1.439{\pm}0.761$ \\
  & FNO      & $0.290{\pm}0.193$ & $0.397{\pm}0.239$ & $0.663{\pm}0.271$ & $0.754{\pm}0.252$ & $0.824{\pm}0.223$ \\
\addlinespace
\multirow{4}{*}{Burgers}
  & NCA      & $\mathbf{0.040{\pm}0.008}$ & $\mathbf{0.048{\pm}0.013}$ & $\mathbf{0.074{\pm}0.032}$ & $\mathbf{0.114{\pm}0.091}$ & $\mathbf{0.196{\pm}0.229}$ \\
  & PDE - Net  & $0.093{\pm}0.015$ & $0.174{\pm}0.043$ & $0.834{\pm}0.460$ & $1.271{\pm}0.872$ & $1.754{\pm}1.543$ \\
  & PINN     & $0.750{\pm}0.176$ & $0.928{\pm}0.269$ & $1.384{\pm}0.612$ & $1.551{\pm}0.751$ & $1.948{\pm}1.244$ \\
  & FNO      & $0.301{\pm}0.192$ & $0.417{\pm}0.232$ & $0.697{\pm}0.245$ & $0.786{\pm}0.219$ & $0.853{\pm}0.186$ \\
\addlinespace
\multirow{4}{*}{Fisher - KPP}
  & NCA      & $\mathbf{0.012{\pm}0.002}$ & $\mathbf{0.021{\pm}0.007}$ & $\mathbf{0.127{\pm}0.065}$ & $\mathbf{0.211{\pm}0.133}$ & $\mathbf{0.315{\pm}0.244}$ \\
  & PDE - Net  & $0.077{\pm}0.013$ & $0.099{\pm}0.027$ & $0.327{\pm}0.154$ & $0.487{\pm}0.265$ & $0.674{\pm}0.420$ \\
  & PINN     & $0.669{\pm}0.124$ & $0.928{\pm}0.234$ & $1.393{\pm}0.572$ & $1.556{\pm}0.785$ & $1.817{\pm}1.166$ \\
  & FNO      & $0.290{\pm}0.193$ & $0.398{\pm}0.238$ & $0.661{\pm}0.274$ & $0.751{\pm}0.259$ & $0.820{\pm}0.235$ \\
  \addlinespace
\multirow{4}{*}{Allen - Cahn}
  & NCA      & $0.008{\pm}0.0002$ & $0.013{\pm}0.0003$ & $0.032{\pm}0.001$ & $0.042{\pm}0.002$ & $0.052{\pm}0.003$ \\
  & PDE - Net  & $\mathbf{0.004{\pm}0.0001}$ & $\mathbf{0.006{\pm}0.0002}$ & $\mathbf{0.010{\pm}0.0005}$ & $\mathbf{0.013{\pm}0.001}$ & $\mathbf{0.015{\pm}0.001}$ \\
  & PINN     & $0.069{\pm}0.004$ & $0.109{\pm}0.007$ & $0.076{\pm}0.008$ & $0.106{\pm}0.006$ & $0.126{\pm}0.006$ \\
  & FNO      & $0.033{\pm}0.002$ & $0.055{\pm}0.003$ & $0.102{\pm}0.006$ & $0.121{\pm}0.008$ & $0.138{\pm}0.010$ \\
\addlinespace
\multirow{4}{*}{Advection}
  & NCA & $\mathbf{0.097{\pm}0.002}$ & $\mathbf{0.184{\pm}0.004}$ & $\mathbf{0.388{\pm}0.008}$ & $0.428{\pm}0.009$ & $0.457{\pm}0.009$ \\
  & PDE - Net  & $0.342{\pm}0.026$ & $0.399{\pm}0.029$ & $0.668{\pm}0.047$ & $0.807{\pm}0.068$ & $0.951{\pm}0.099$ \\
  & PINN     & $0.993{\pm}0.023$ & $1.023{\pm}0.026$ & $0.975{\pm}0.036$ & $0.924{\pm}0.029$ & $0.878{\pm}0.019$ \\
  & FNO      & $0.343{\pm}0.044$ & $0.362{\pm}0.054$ & $0.407{\pm}0.073$ & $\mathbf{0.424{\pm}0.079}$ & $\mathbf{0.437{\pm}0.084}$ \\
\bottomrule
\end{tabular}%
}
\end{table}

Table~\ref{tab:full} reports mean relative $\ell_2$ error for each temporal horizon for every PDE and model. 
For the heat equation, NCA achieves the lowest error at every reported horizon, with the advantage becoming evident at long horizons.
At $T=1000$, which is twice the training temporal horizon, its relative error is $0.561$, compared with $0.709$ for PDE - Net, $0.824$ for FNO, and $1.439$ for PINN. 
Thus, relative to the next-best baseline, PDE - Net, NCA reduces the error at $T=1000$ by approximately $21\%$. 
The same trend can be found for Burgers and Fisher - KPP. 
At $T=1000$, NCA obtains errors of $0.196$ and $0.315$, respectively, whereas PDE - Net reaches $1.754$ and $0.674$. This corresponds to reductions of approximately $89\%$ and $53\%$ relative to PDE - Net on these two equations.

On Burgers, the NCA errors remain below $0.20$ throughout the entire evaluated interval, increasing from $0.040$ at $T=100$ to $0.196$ at $T=1000$. 
In contrast, PDE - Net grows from $0.093$ to $1.754$, while FNO grows from $0.301$ to $0.853$. 
Since the objective of the experiment is not merely to obtain a good one-step or short-horizon prediction, but to maintain a stable approximation under repeated application of the learned update rule, these results are particularly important.
The widening gap between NCA and the baselines indicates that the learned local dynamics of NCA are more robust to other well-known baselines.
A similar pattern is also observed for Fisher - KPP. NCA has the lowest error at all five horizons, starting at $0.012$ at $T=100$ and reaching $0.315$ at $T=1000$. 
PDE - Net remains the closest baseline but its error grows to $0.674$, while FNO and PINN reach $0.820$ and $1.817$, respectively. 
The large error of PINN at long horizons is consistent with the difficulty of maintaining accurate trajectory predictions when a model is optimized primarily through the PDE residual and boundary/initial-condition constraints.

For Allen - Cahn, however, PDE - Net is consistently more accurate, achieving errors of $0.004$, $0.006$, $0.010$, $0.013$, and $0.015$ across the five horizons, compared with $0.008$, $0.013$, $0.032$, $0.042$, and $0.052$ for NCA. 
Regardless, NCA still maintains a relatively small absolute error throughout the rollout, and its $T=1000$ error remains substantially below those of PINN ($0.126$) and FNO ($0.138$). 
The result suggests that PDE - Net is especially well matched to the dynamics of the Allen - Cahn configuration used here, while NCA provides a more consistently strong performance across the broader collection of PDEs. 
\begin{figure}
    \centering
    \includegraphics[width=0.7\linewidth]{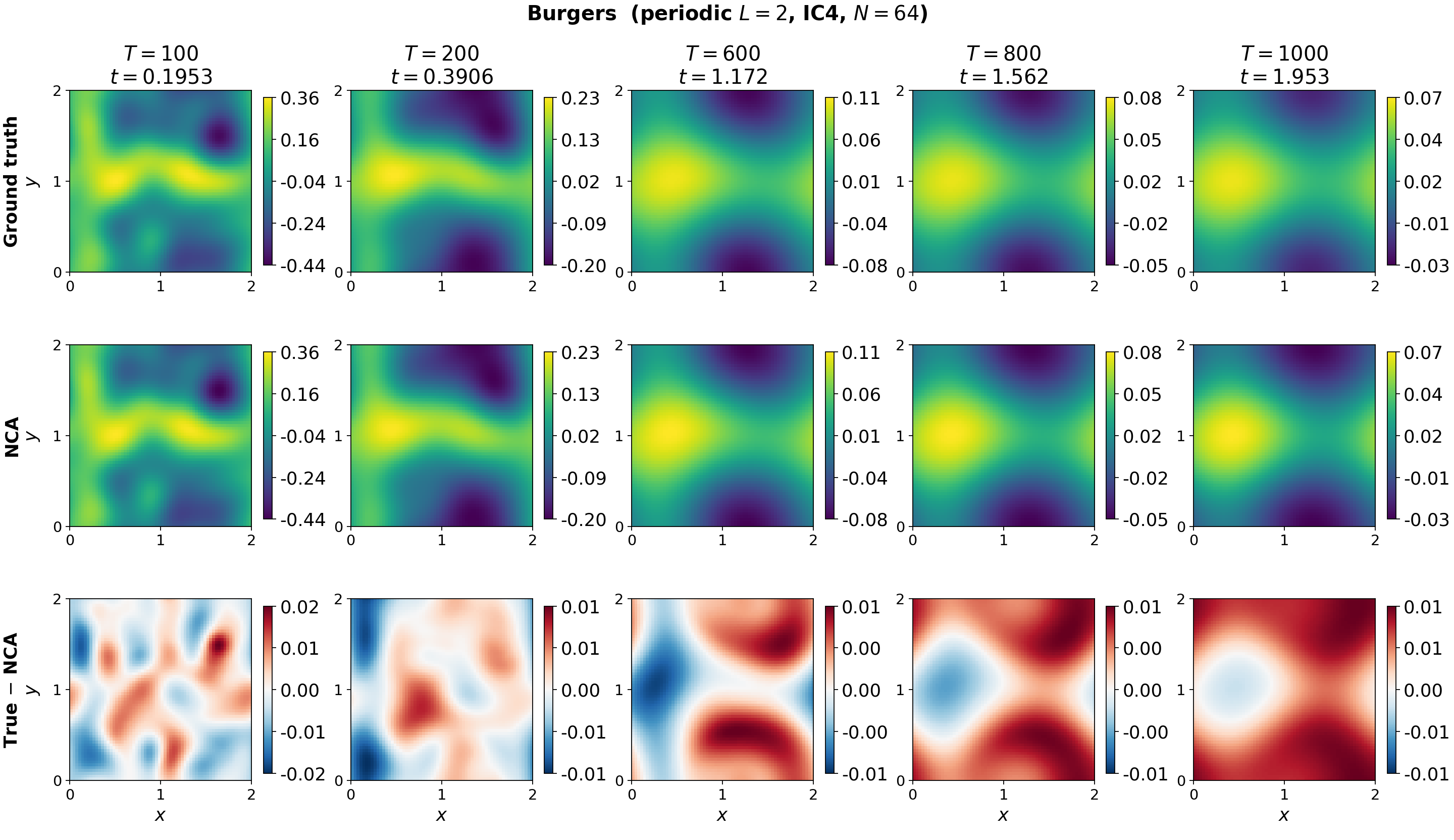}
    \caption{A sample true and learned solution of the Burgers equation using NCA, along with the difference of true solution and learned solution.}
    \label{fig:burgers}
\end{figure}
The behavior of PINN is particularly notable across the benchmark. On heat, Burgers, and Fisher - KPP, its relative error exceeds $1.0$ before or by the longest rollout horizon, reaching $1.439$, $1.948$, and $1.817$, respectively, at $T=1000$. These values indicate that the accumulated rollout error becomes comparable to or larger than the magnitude of the reference solution. PINN performs better on Allen--Cahn, where its $T=1000$ error is $0.126$, but it remains worse than both NCA and PDE - Net. This consistent degradation at long horizons highlights the difference between learning a solution satisfying the governing equation and learning a stable discrete evolution operator that can be repeatedly applied.

FNO occupies an intermediate position in most of the experiments. It is substantially more accurate than PINN on heat, Burgers, and Fisher - KPP, but remains behind NCA at long horizons. 
For example, at $T=1000$, NCA improves over FNO by approximately $32\%$ on heat, $77\%$ on Burgers, and $62\%$ on Fisher - KPP. 
FNO is the strongest model on the advection experiment at the longest horizon, obtaining a relative error of $0.437$ compared with $0.457$ for the NCA. 

\begin{figure}
    \centering
    \includegraphics[width=0.4\linewidth]{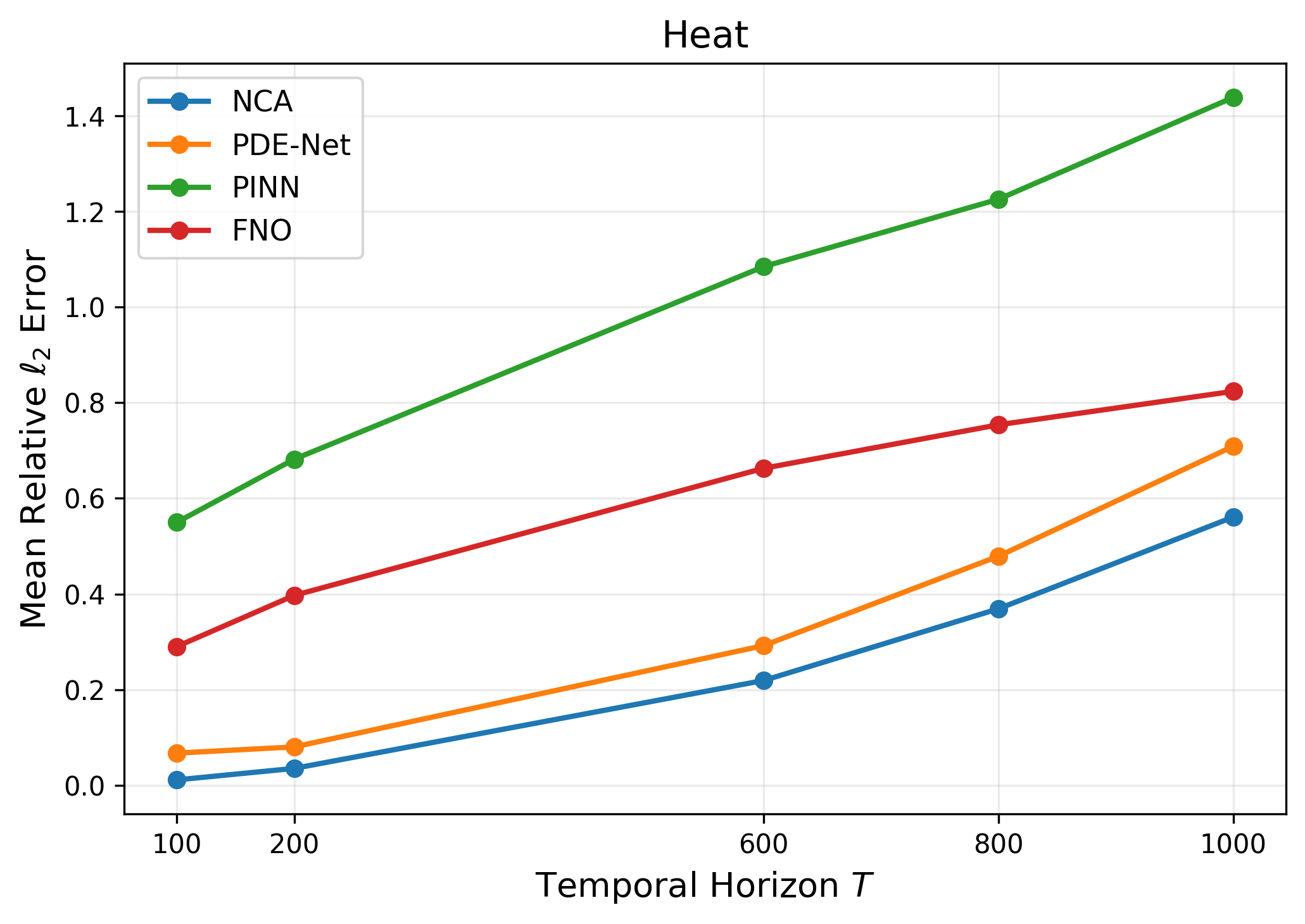} 
\includegraphics[width=0.4\linewidth]{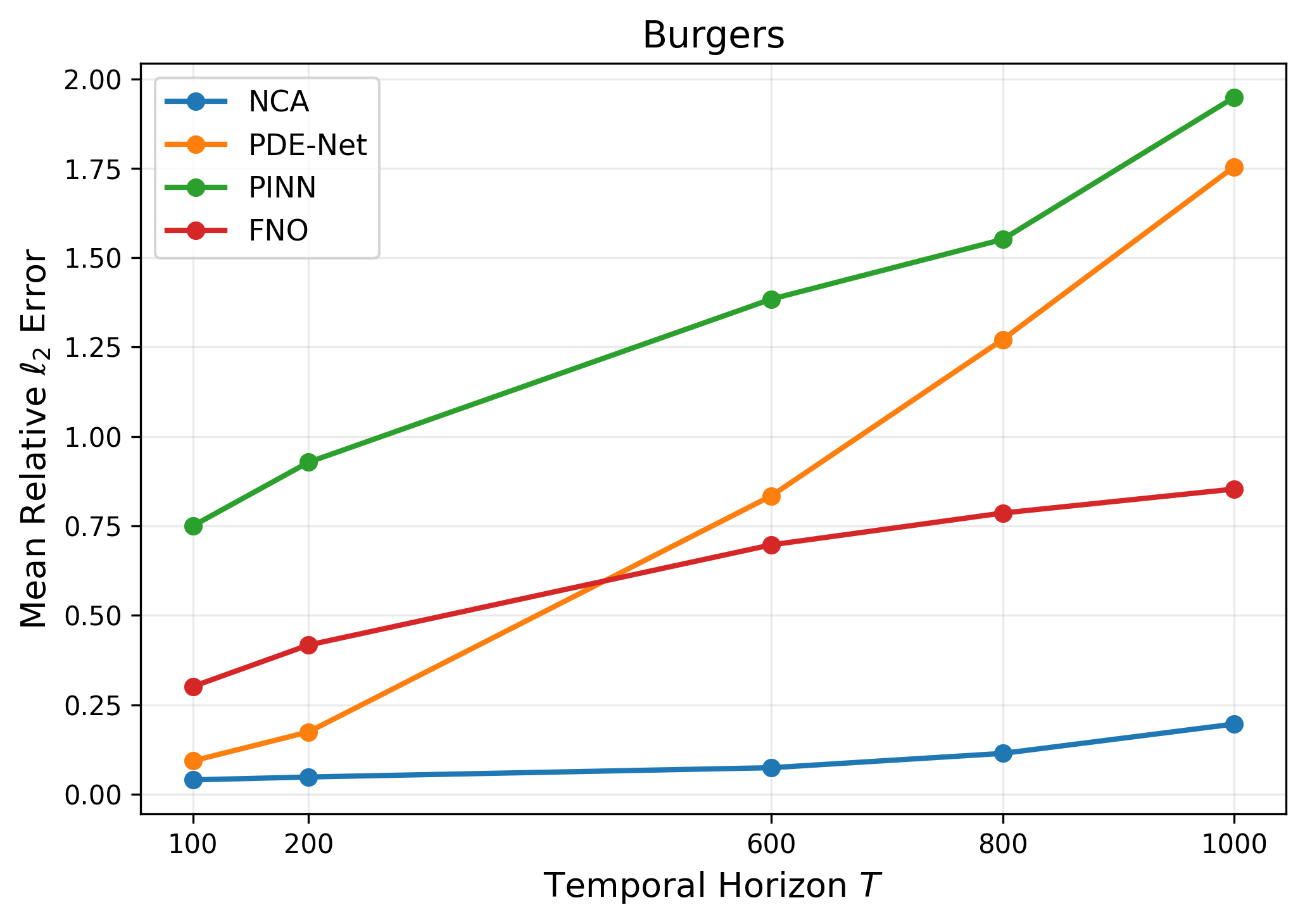}\\
\includegraphics[width=0.4\linewidth]{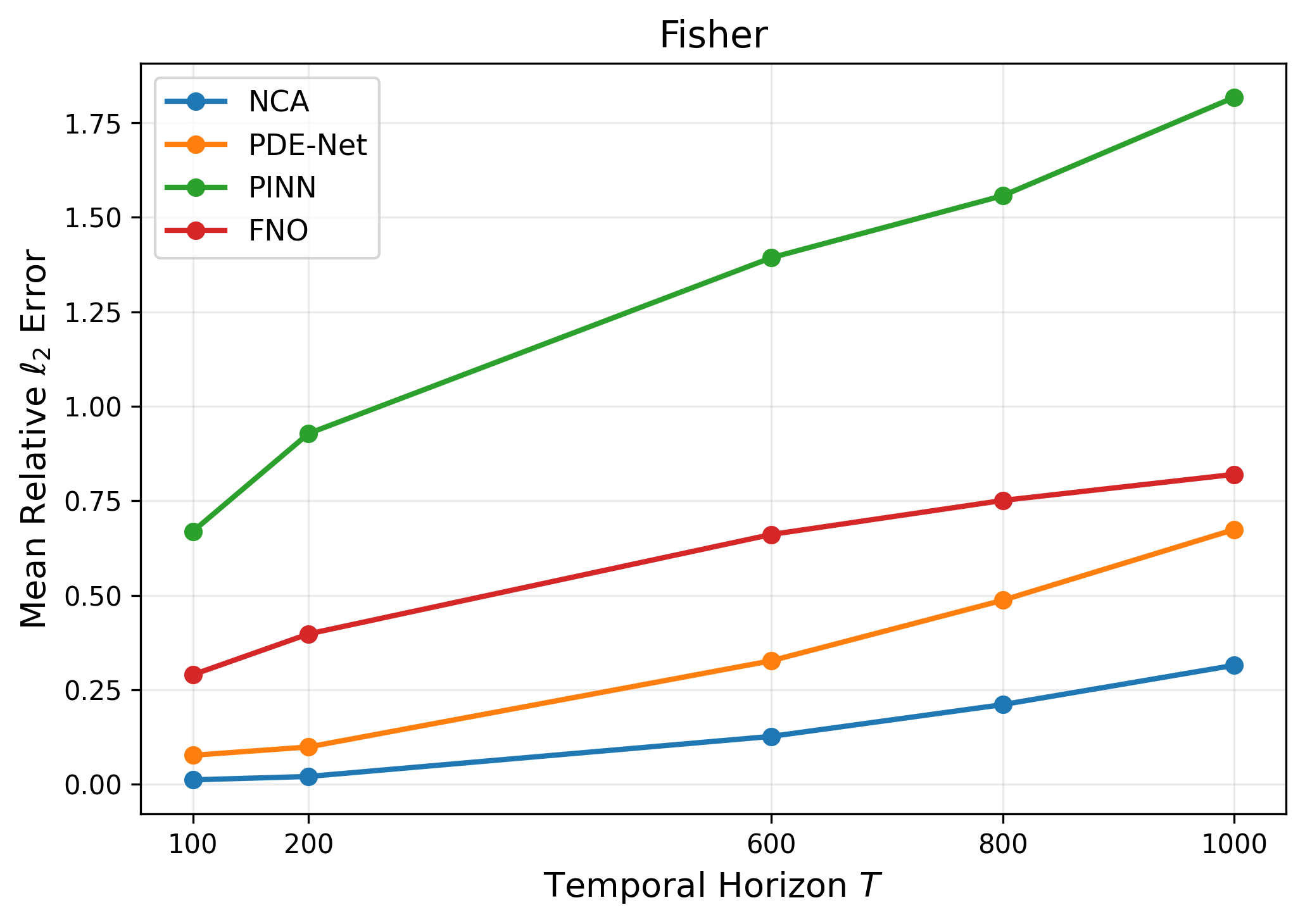} 
\includegraphics[width=0.4\linewidth]{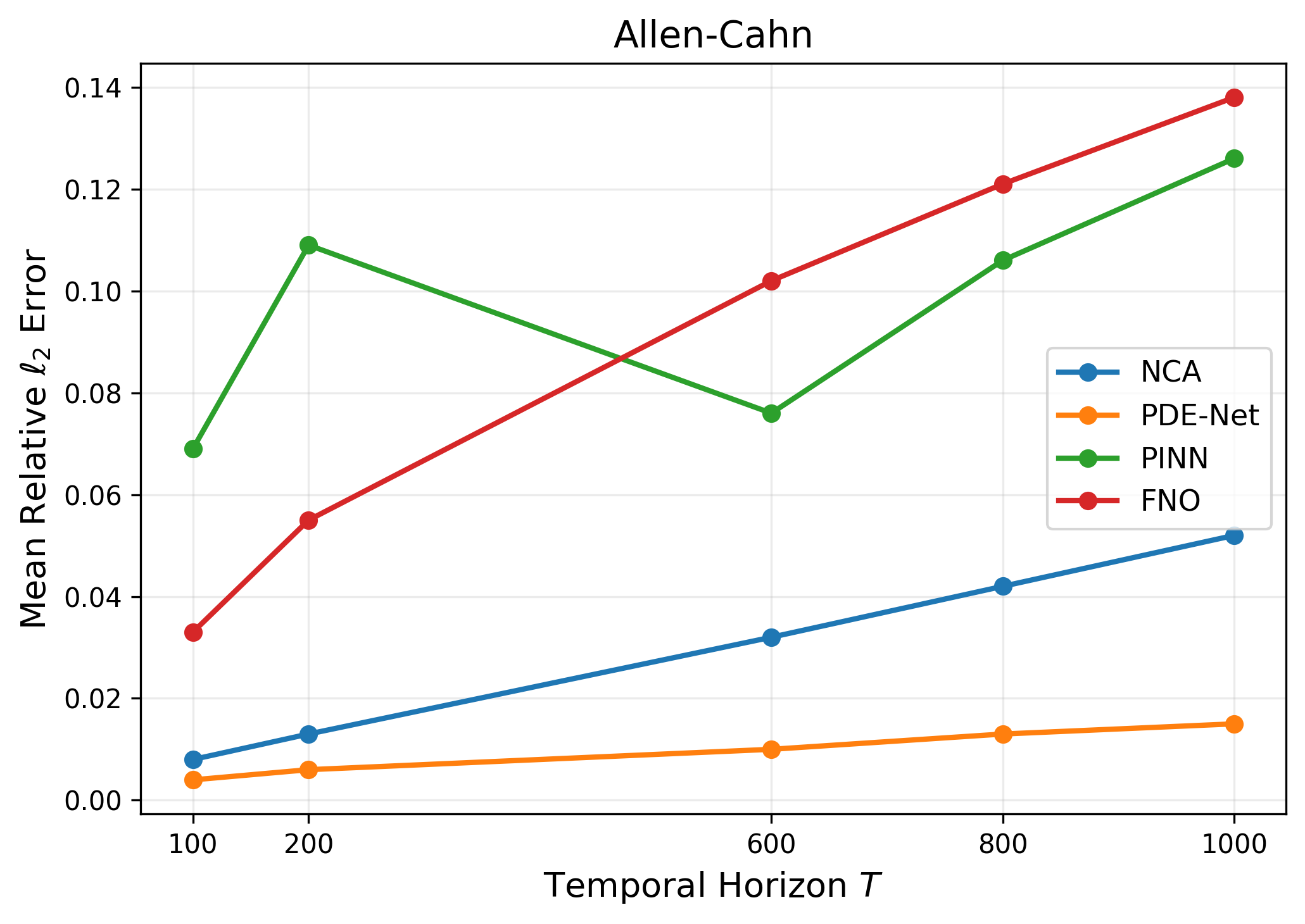}\\
\includegraphics[width=0.4\linewidth]{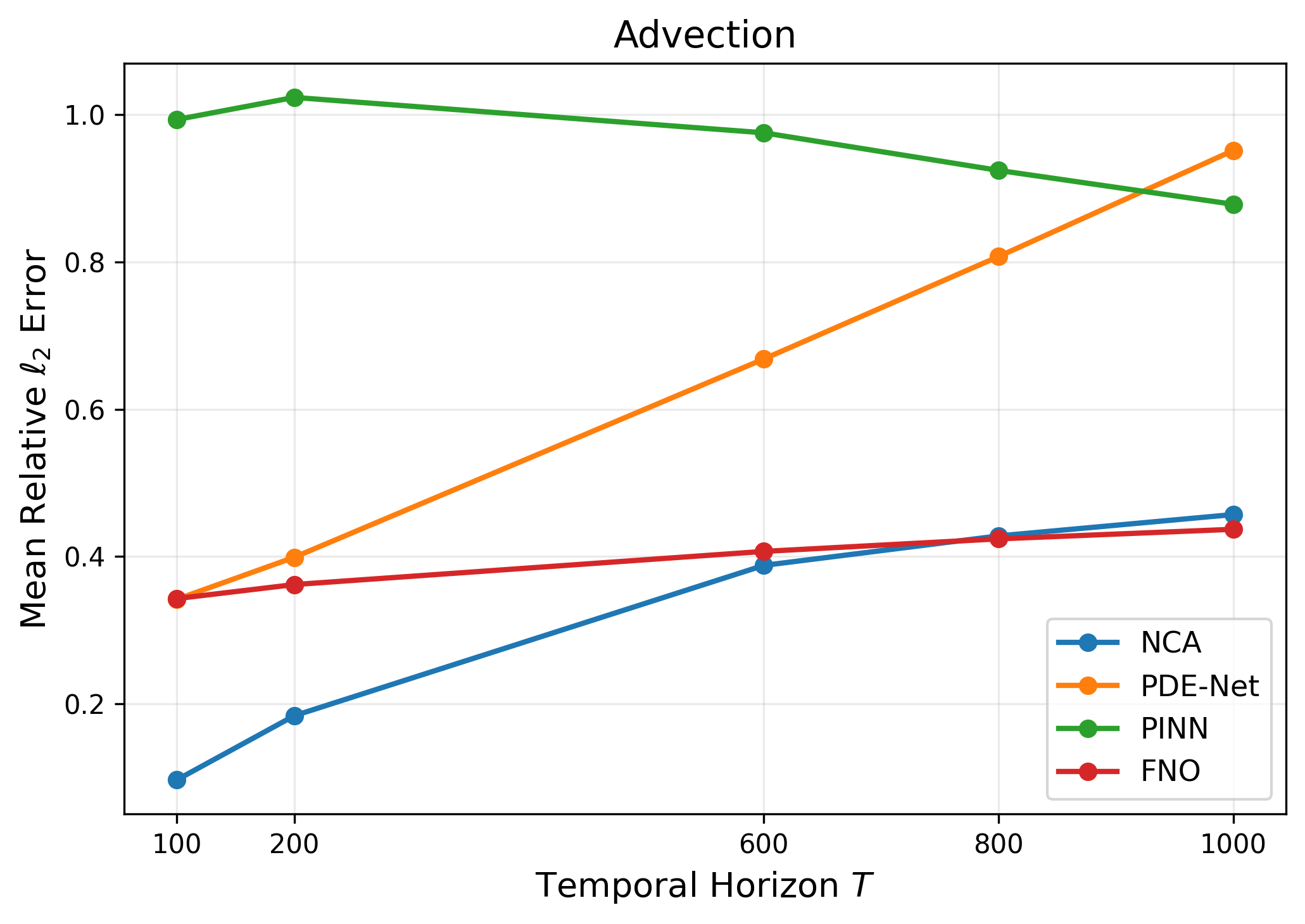}
    \caption{Error growth across temporal horizon for different methods.}
    \label{fig:error}
\end{figure}

An additional observation from Table~\ref{tab:full} is that the performance is more pronounced at longer horizons than at shorter ones. At $T=100$, several baselines still produce moderate errors, whereas repeated application of their learned dynamics causes the error to accumulate rapidly. 
NCA, in contrast, maintains comparatively gradual error growth on Burgers and Fisher - KPP and remains the best-performing model on heat throughout the evaluated interval. 
This distinction between short- and long-horizon behavior is central to the proposed formulation: a model can achieve a reasonable approximation at an early horizon while still possessing a learned update rule that is unstable or poorly conditioned under repeated composition. 
The long-horizon results therefore provide a more stringent test of whether the learned operator captures the underlying temporal dynamics rather than simply interpolating the training trajectories.
In order to visualize better, we have also plotted the results in Table \ref{tab:full} and Figure \ref{fig:error}. 
The plots show that relative error generally increases with temporal horizon for all methods, but NCA exhibits the most stable long-horizon behavior on the majority of the PDEs, where its error remains consistently below or close to the competing approaches and grows relatively gradually as $T$ increases. 

Taken together, Table~\ref{tab:full} supports three conclusions. 
\begin{enumerate}
    \item NCA provides the strongest long-horizon performance on majority of the benchmark PDEs.
    \item Its advantage becomes particularly large on Burgers and Fisher - KPP, where the competing models accumulate substantial rollout error.
    \item The Allen--Cahn result demonstrates that NCA is not uniformly superior and that the structure of the underlying PDE and initial condition matters.
\end{enumerate}

\subsection{Robustness to Noise}
\label{sec:noise}

To test whether the model learning is susceptible to
observation noise, we retrained every model with additive Gaussian noise
on the training spatiotemporal snapshots,
\[
u_{\mathrm{noisy}} = u + \sigma\,\mathrm{std}(u)\,\varepsilon,
\quad \varepsilon\sim\mathcal{N}(0,1),
\]
at relative levels $\sigma\in\{0.05,0.10\}$ (5\%, 10\%),
while keeping evaluation on the clean solutions.

\begin{table}[h!]
\centering
\small
\caption{Mean relative $\ell_2$ error under additive Gaussian training noise of varying strengths.}
\label{tab:noise}
\resizebox{\textwidth}{!}{%
\begin{tabular}{llcccccc}
\toprule
PDE & Model & \multicolumn{2}{c}{$\sigma{=}0$} & \multicolumn{2}{c}{$\sigma{=}5\%$} & \multicolumn{2}{c}{$\sigma{=}10\%$} \\
 & & $T{=}100$ & $T{=}1000$ & $T{=}100$ & $T{=}1000$ & $T{=}100$ & $T{=}1000$ \\
\midrule
\multirow{4}{*}{Heat} & NCA & $\mathbf{0.011{\pm}0.002}$ & $\mathbf{0.561{\pm}0.486}$ & $\mathbf{0.017{\pm}0.003}$ & $\mathbf{0.246{\pm}0.184}$ & $\mathbf{0.018{\pm}0.004}$ & $\mathbf{0.271{\pm}0.202}$ \\
 & PDE-Net & $0.067{\pm}0.012$ & $0.709{\pm}0.493$ & $0.125{\pm}0.015$ & $2.309{\pm}1.373$ & $0.129{\pm}0.024$ & $1.701{\pm}1.016$ \\
 & FNO & $0.290{\pm}0.193$ & $0.824{\pm}0.223$ & $0.300{\pm}0.207$ & $0.820{\pm}0.244$ & $0.300{\pm}0.207$ & $0.820{\pm}0.244$ \\
 & PINN & $0.550{\pm}0.082$ & $1.439{\pm}0.761$ & $0.574{\pm}0.119$ & $1.428{\pm}0.577$ & $0.572{\pm}0.080$ & $1.907{\pm}0.961$ \\
\addlinespace
\multirow{4}{*}{Burgers} & NCA & $\mathbf{0.040{\pm}0.008}$ & $\mathbf{0.196{\pm}0.229}$ & $\mathbf{0.050{\pm}0.009}$ & $\mathbf{0.725{\pm}0.371}$ & $\mathbf{0.049{\pm}0.010}$ & $\mathbf{0.598{\pm}0.300}$ \\
 & PDE-Net & $0.093{\pm}0.015$ & $1.754{\pm}1.543$ & $0.094{\pm}0.018$ & $1.632{\pm}1.227$ & $0.116{\pm}0.025$ & $2.053{\pm}1.268$ \\
 & FNO & $0.301{\pm}0.192$ & $0.853{\pm}0.186$ & $0.306{\pm}0.209$ & $0.841{\pm}0.213$ & $0.306{\pm}0.209$ & $0.841{\pm}0.213$ \\
 & PINN & $0.750{\pm}0.176$ & $1.948{\pm}1.244$ & $0.772{\pm}0.131$ & $1.813{\pm}0.876$ & $0.640{\pm}0.104$ & $2.065{\pm}1.042$ \\
\addlinespace
\multirow{4}{*}{Fisher} & NCA & $\mathbf{0.012{\pm}0.002}$ & $\mathbf{0.315{\pm}0.244}$ & $\mathbf{0.011{\pm}0.002}$ & $\mathbf{0.123{\pm}0.074}$ & $\mathbf{0.013{\pm}0.002}$ & $\mathbf{0.107{\pm}0.034}$ \\
 & PDE-Net & $0.077{\pm}0.013$ & $0.674{\pm}0.420$ & $0.099{\pm}0.013$ & $1.927{\pm}1.301$ & $0.094{\pm}0.015$ & $1.183{\pm}0.526$ \\
 & FNO & $0.290{\pm}0.193$ & $0.820{\pm}0.235$ & $0.300{\pm}0.204$ & $0.831{\pm}0.220$ & $0.300{\pm}0.204$ & $0.831{\pm}0.220$ \\
 & PINN & $0.669{\pm}0.124$ & $1.817{\pm}1.166$ & $0.418{\pm}0.060$ & $1.843{\pm}0.957$ & $0.470{\pm}0.079$ & $1.751{\pm}0.925$ \\
\addlinespace
\multirow{4}{*}{Allen--Cahn} & NCA & $0.008{\pm}0.0002$ & $0.052{\pm}0.003$ & $0.005{\pm}0.000$ & $\mathbf{0.011{\pm}0.001}$ & $0.005{\pm}0.000$ & $0.014{\pm}0.001$ \\
 & PDE-Net & $\mathbf{0.004{\pm}0.0001}$ & $\mathbf{0.015{\pm}0.001}$ & $\mathbf{0.005{\pm}0.000}$ & $0.019{\pm}0.002$ & $\mathbf{0.004{\pm}0.000}$ & $\mathbf{0.009{\pm}0.001}$ \\
 & FNO & $0.033{\pm}0.002$ & $0.138{\pm}0.010$ & $0.034{\pm}0.002$ & $0.139{\pm}0.012$ & $0.034{\pm}0.002$ & $0.139{\pm}0.012$ \\
 & PINN & $0.069{\pm}0.004$ & $0.126{\pm}0.006$ & $0.059{\pm}0.004$ & $0.083{\pm}0.005$ & $0.072{\pm}0.005$ & $0.093{\pm}0.005$ \\
\addlinespace
\multirow{4}{*}{Advection} & NCA & $\mathbf{0.097{\pm}0.002}$ & $0.457{\pm}0.009$ & $0.781{\pm}0.035$ & $0.958{\pm}0.050$ & $0.781{\pm}0.035$ & $0.957{\pm}0.050$ \\
 & PDE-Net & $0.342{\pm}0.026$ & $0.951{\pm}0.099$ & $\mathbf{0.322{\pm}0.030}$ & $0.633{\pm}0.091$ & $\mathbf{0.324{\pm}0.030}$ & $\mathbf{0.480{\pm}0.057}$ \\
 & FNO & $0.343{\pm}0.044$ & $\mathbf{0.437{\pm}0.084}$ & $0.336{\pm}0.052$ & $\mathbf{0.415{\pm}0.103}$ & $0.336{\pm}0.052$ & $\mathbf{0.415{\pm}0.103}$ \\
 & PINN & $0.993{\pm}0.023$ & $0.878{\pm}0.019$ & $1.016{\pm}0.024$ & $0.893{\pm}0.051$ & $1.010{\pm}0.026$ & $0.861{\pm}0.025$ \\
\bottomrule
\end{tabular}
}
\end{table}

The results in Table~\ref{tab:noise} show that NCA exhibits good robustness to moderate levels of additive Gaussian noise in the training data, although the effect varies across PDEs.
For the heat and Fisher - KPP equations, NCA maintains relatively low errors even at 5\% and 10\% noise, with substantially smaller errors at $T{=}1000$ than PDE-Net, FNO, and PINN.
On both PDEs the long-horizon error \emph{decreases} relative to the clean run
(heat: $0.561$ to $0.246$ at $5\%$ noise level; Fisher: $0.315$ to $0.123$).
This robustness may arise from the local fixed-stencil perception used by NCA, which restricts the information passed to the update rule to structured local spatial features and may therefore reduce the influence of high-frequency perturbations introduced by noisy targets.
In contrast, PDEs with stronger transport or nonlinear dynamics can be more
sensitive to perturbations because small local errors may be amplified during repeated rollouts.
This is clearest for advection, where NCA's error increases substantially under noisy training, while FNO and PDE-Net remain far more stable.
Burgers exhibits an intermediate behavior: although its error increases with noise ($0.196$ to $0.60$ - $0.73$ at $T{=}1000$), NCA still retains a lower long-horizon error than PDE-Net, FNO, and PINN.
For the Allen - Cahn PDE, at $5\%$ noise NCA attains the lowest error for $T{=}1000$ error ($0.011$ vs.\ $0.019$), while PDE - Net has lowest errors at $10\%$ noise ($0.009$ vs.\ $0.014$), suggesting that the inductive bias of the NCA architecture is not equally
advantageous for all dynamics or noise levels.
Overall, these results suggest that NCA's local, fixed-stencil perception and gated updates can provide meaningful robustness to noisy training data, particularly for diffusion- and reaction-dominated dynamics, while transport-dominated equations remain more sensitive to perturbations
accumulated during long rollouts.

\subsection{Interpreting Learned Perception Gates}
\label{sec:interpretability}

Each NCA channel multiplies the inputs with four fixed finite difference matrices representing identity, $\partial_x$, $\partial_y$, and the Laplacian.
Trainable scalar \emph{gates} multiply with each of these four outputs, together with an $\ell_1$ sparsity penalty
($\lambda{=}10^{-4}$) which act as data-driven operator selectors.
Table~\ref{tab:nca-gates} reports the gate magnitude associated with each stencil.
Larger magnitudes indicate a preference for the corresponding fixed operator among those available in the perception layer.
Values near zero indicate that the corresponding stencil has less importance under the $\ell_1$ penalty.

\begin{table}[ht]
\centering
\small
\caption{NCA learned perception-gate magnitudes. The top two dominant terms are compared with the true operators in the equations. Overlapping operators are highlighted in bold.}
\label{tab:nca-gates}
\begin{tabular}{lccccrr}
\toprule
PDE & Identity ($I$) & $\partial_x$ & $\partial_y$ & Laplace ($\Delta$) & Dominant & True \\
\midrule
Heat & 0.0206 & $0.00e+00$ & $0.00e+00$ & 0.0368 & I, $\boldsymbol{\Delta}$ & $\Delta$ \\
Burgers & 0.0232 & 0.0081 & $0.00e+00$ & 0.0374 & $\Delta$, $\mathbf{I}$ & $\partial_x$, $\mathbf{I} $ \\
Fisher - KPP & 0.0205 & $0.00e+00$ & $0.00e+00$ & 0.0370 & $\boldsymbol{I, \Delta}$ & $\Delta$, I \\
Allen Cahn & 0.0167 & 0.0019 & 0.0016 & 0.0062 & $\boldsymbol{\Delta, I}$ & $\Delta$, I \\
Advection & 0.0109 & 0.1067 & 0.0001 & 0.0194 & $\boldsymbol{\partial_x}$, $\Delta$ & $\partial_x$ \\
\bottomrule
\end{tabular}
\end{table}

On the heat, Allen - Cahn and Fisher - KPP PDEs, $\partial_x$ and $\partial_y$ gates have the lowest values while the Laplacian gate is largest, followed by the identity channels.  
This matches closely with the true equations, which
are $\Delta$ dominated.  
Burgers follows the same Laplacian-first pattern but with a relatively larger $\partial_x$ gate, consistent with the viscous Burgers combining diffusion and a state-dependent advective flux. 
The model therefore allocates weight to both the
Laplacian and the horizontal-derivative stencil, while $\partial_y$ remains inactive.
Advection gates allocate the largest entry to $\partial_x$, which agrees with the true dominating term in the original PDE.
Across the five benchmark PDEs where all models share a comparable implementation, the learned gates are sparse and PDE-aligned.
This supports the discussion in Section~\ref{sec:discussion} that NCA's fixed, sparse perception filters can potentially learn physically interpretable updates rather than an unconstrained black-box map.

\section{Discussion}
\label{sec:discussion}

NCA is trained to learn the updates of the next timestep by considering its immediate neighbouring cells. 
NCA is trained as a one-step map (truncation $K=1$) on $R=500$ closed-loop steps, then evaluated by composing that map to $T=1000$.
This trains the model on states encountered during closed-loop evolution, while the evaluation directly measures the stability of the resulting update rule over long rollouts.
The perception layer uses a dictionary of various finite-difference stencils such as identity, Sobel, Laplacian operators which help in identifying the operators/PDE terms which may be present in the equation. We discuss the results of each of the baseline methods in detail below.

The PDE - Net is trained using learnable $5\times5$ filters with a pointwise MLP.
The performance is behind NCA on the heat equation at $T=100$ (relative error of 0.067 for PDE - Net vs $0.011$ for NCA).
The gap widens substantially by the time we reach $T=1000$ ($0.709$ for PDE - Net vs.\ $0.561$ for NCA).
However, PDE - Net outperforms NCA at every horizon for the Allen - Cahn PDE. 
On Burgers and Fisher - KPP, PDE - Net's error grows sharply beyond $T=200$ (e.g., $0.174$ to $1.754$ on Burgers between $T=200$ and $T=1000$). 

The PINN baseline reported in Table~\ref{tab:full} uses a CNN with initial condition
$u(x,y,0)=u_0(x,y)$ and a physics residual. 
The PINN learns
$u_0 \mapsto u(\cdot,T)$ for a random query horizon $T$, whereas
NCA/PDE - Net/FNO learn the Markov transition $u^n \mapsto u^{n+1}$ and
compose it $T$ times. Long-horizon composition is never directly enforced
during PINN training, leading to an underperformance beyond the training temporal horizon.
Even with normalized residuals, the
PINN must simultaneously satisfy the PDE, the initial condition, and data. 

 On heat, Burgers, and Fisher - KPP its performance in terms of relative errors is between PDE - Net and PINN. 
At $T=1000$ it is substantially behind NCA.
On Allen - Cahn it is close to PDE - Net but beats PINN. 

\subsection{Ablation Study}
In this section we provide a controlled ablation on the
\textbf{heat} and \textbf{Burgers} equations with a random Gaussian initial condition, $L=2$, and evaluate it for  $T\in\{100,200,600,800,1000\}$. We compare the average full field relative errors of NCA, PINNs and FNO for using different numbers of trainable parameters and the training and evaluation times.
We consider two scenarios for model comparisons: the original version whose results were discussed in Section \ref{sec:results} and a \textit{par-matched} version where the number of trainable parameters of NCA, FNO and PINNs have been adjusted so they remain fairly similar. 
In the original version, we consider the best versions of the model i.e., the number of trainable parameters for FNO and PINN are relatively high in comparison to NCA.
For the \textit{par-matched} NCA version, we use at least 64 channels with minimum 128 hidden neurons. 

\begin{table}[t]
\centering
\small
\caption{Comparison of baselines and training times under their best performance along with the number of trainable parameters matched with NCA. ``Train~(s)'' is single-run wall-clock time on one NVIDIA H100 GPU. The times cover the fixed training loop (Adam updates plus online
finite-difference reference steps) and exclude hyperparameter search and trajectory sampling.}
\label{tab:ablation-n64}
\begin{tabular}{lllcccc}
\toprule
PDE & Protocol & Model & Params & Train (s) & $T = 100$ & $T = 1000$ \\
\midrule
\multirow{8}{*}{Heat}
  & original & NCA     & 5.3k   & 696   & 0.016 & \textbf{0.139} \\
  & original & PDE - Net & 5.1k   & 957   & 0.080 & 0.707 \\
  & original & FNO     & 2.36M  & 6906  & 0.314 & 1.026 \\
  & original & PINN    & 208k   & 195   & 0.455 & 1.468 \\
  & par-matched     & NCA     & 61.9k  & 2295  & 0.010 & \textbf{0.179} \\
  & par-matched     & PDE - Net & 59.0k  & 3052  & 0.268 & 0.979 \\
  & par-matched     & FNO     & 65.9k  & 23063 & 0.307 & 1.231 \\
  & par-matched     & PINN    & 52.9k  & 380   & 0.544 & 1.430 \\
\addlinespace
\multirow{8}{*}{Burgers}
  & original & NCA     & 5.3k   & 871   & 0.030 & 0.625 \\
  & original & PDE - Net & 5.1k   & 1161  & 0.087 & \textbf{0.580} \\
  & original & FNO     & 2.36M  & 7099  & 0.315 & 1.015 \\
  & original & PINN    & 208k   & 442   & 0.484 & 1.833 \\
  & par-matched     & NCA     & 61.9k  & 2917  & 0.036 & \textbf{0.256} \\
  & par-matched     & PDE - Net & 59.0k  & 3660  & 0.050 & 0.407 \\
  & par-matched     & FNO     & 65.9k  & 23283 & 0.314 & 1.602 \\
  & par-matched     & PINN    & 52.9k  & 886   & 0.536 & 2.058 \\
\bottomrule
\end{tabular}
\end{table}

\begin{figure}[h!]
\centering
\begin{subfigure}[t]{0.48\textwidth}
  \centering
  \includegraphics[width=\linewidth]{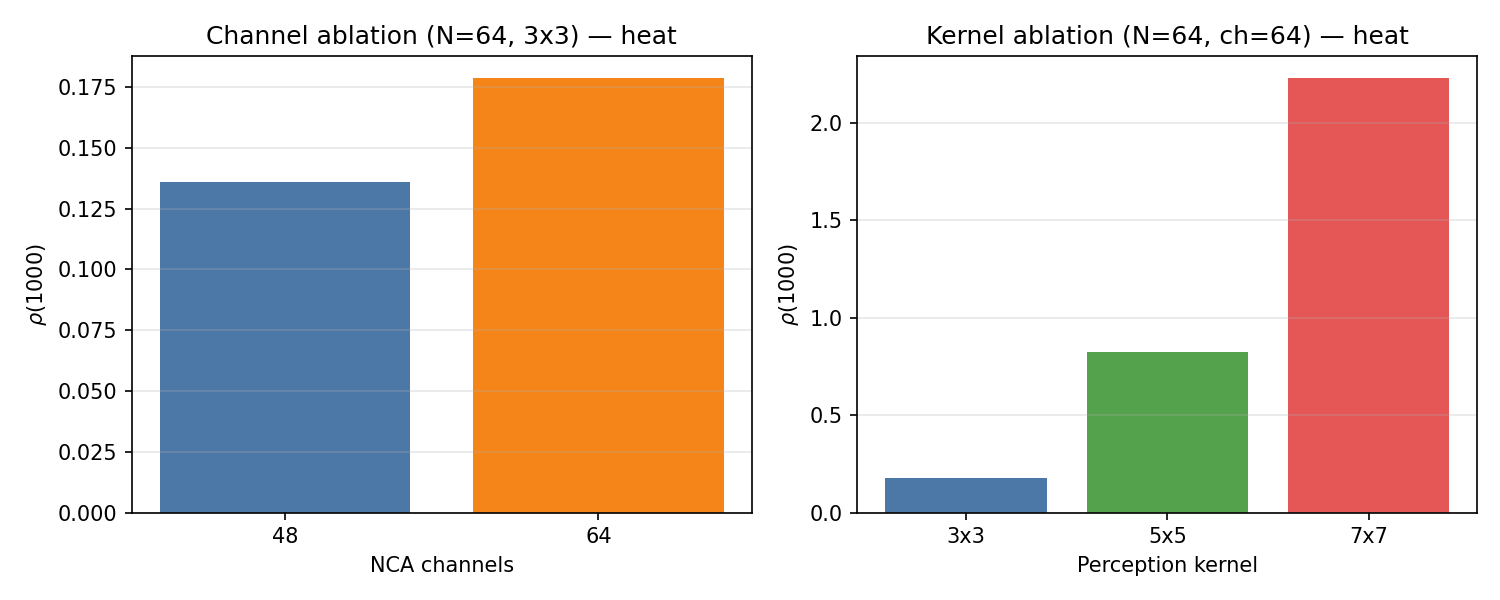}
  \caption{Heat ($N=64$)}
\end{subfigure}\hfill
\begin{subfigure}[t]{0.48\textwidth}
  \centering
  \includegraphics[width=\linewidth]{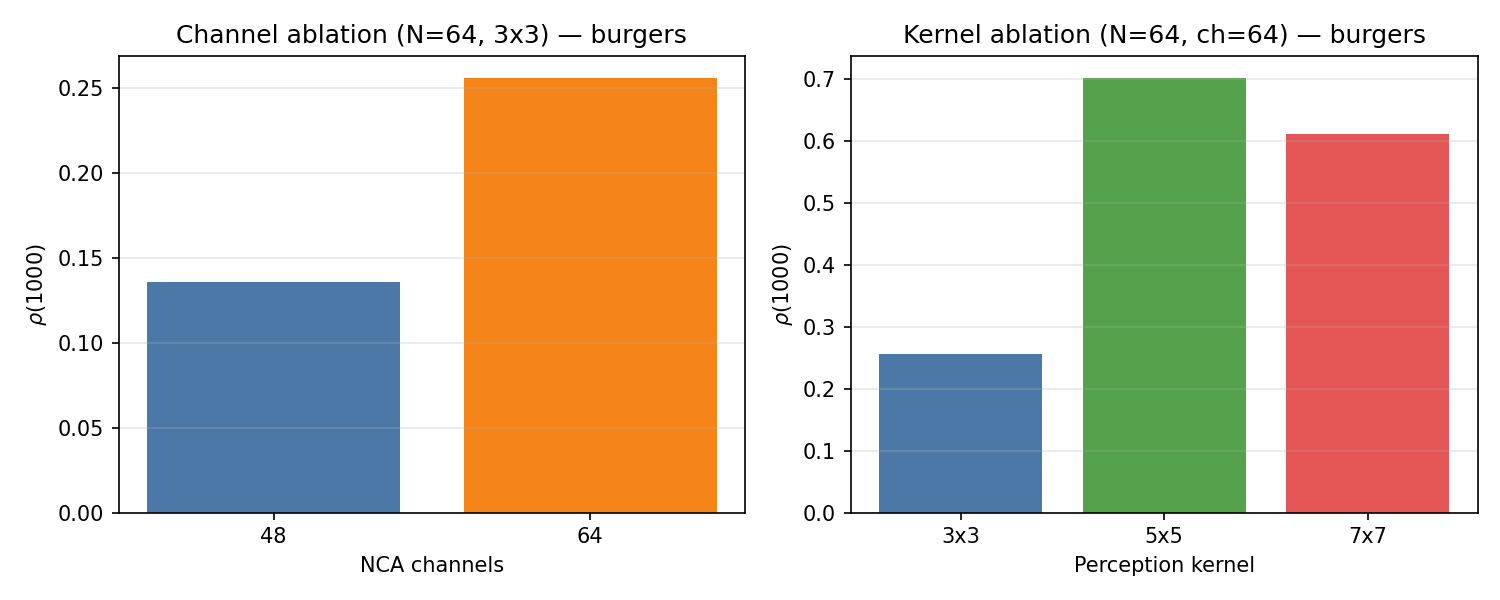}
  \caption{Burgers ($N=64$)}
\end{subfigure}
\caption{NCA channel and perception-kernel ablations at
$N=64$. Compact $3\times3$ kernels dominate wider
$5\times5$ and $7\times7$ alternatives at long horizon.}
\label{fig:ablation-ck}
\end{figure}

\paragraph{Training Time.} Here we will be comparing the long rollout relative errors at $T=1000$ as the resolution changes.
In the fair versions, as $N$ increases, the performance of NCA improves as well
(see Table~\ref{tab:ablation-n64}). Additional simulations on Burgers equation indicated that the NCA error decreases monotonically from $0.820$ to $0.256$ as we changed $N=32$ to $N=64$.
On heat, while errors at $N=32$ are already low at 0.283, the best configuration at $N=64$ gives the lowest error of $0.179$. 
FNO improves with resolution but does not improve enough to be close to NCA, both at the original setting as well as the \textit{par-matched} setting. 
When we reduce the parameters of FNO (and increase parameters of NCA) to match closely with NCA, training FNO is almost 10$\times$ slower than NCA
(e.g., training time of 23000 secs for FNO vs 2300 secs for NCA on heat at $N=64$).
When only considering the training time, PINN trains fastest with about 380 secs but its poor performance for longer temporal horizons makes it a costly tradeoff for the short training time. 

\paragraph{Number of Neighbours for NCA.}
Our proposed NCA uses a 3×3 fixed-stencil perception kernel with trainable scalar gates. We explore if it would also be useful to consider farther neighbours through a $5\times5$ or $7\times7$ kernel. Figure~\ref{fig:ablation-ck} shows that widening the NCA perception kernel hurts long-horizon accuracy where at $N=64$ on heat, relative errors at $T=1000$ rise from $0.179$ for a $3\times3$ kernel to $0.825$ for $5\times5$ kernel and $2.231$ for $7\times7$ kernel.
The same pattern holds on Burgers.
This pattern indicates that for a cell to learn its next temporal update, only the immediate neighbours provide sufficient information. Under the tested architecture and training protocol, the 3$\times$3 perception field provides substantially better long-horizon accuracy than the 5$\times$5 and 7$\times$7 alternatives.
\label{sec:ablation-setup}
\label{sec:orig}
\label{sec:fair}
\label{sec:ablation-full}
\label{sec:ablation-discussion}

\section{Conclusions}
\label{sec:conclusion}
In this paper, we proposed a Neural Cellular Automata (NCA)-based approach for learning the time evolution of solutions to partial differential equations (PDEs). 
The goal was to develop a simple local model that can predict the solution over many time steps while maintaining good accuracy and stability. 
Comparisons of NCA with PDE - Net, FNO, and a PINN baseline using identical settings indicated that at long time horizons, NCA outperformed the other baselines for a majority of the benchmark PDEs. 
The results are consistent with the fact that small errors can grow slowly when the learned update is stable but can grow rapidly when it is not. 
The ablation experiments further showed that NCA performed better on heat and Burgers under different resolutions and is optimal only when learning is considered from immediate neighbours.

\paragraph{Limitations and Future Work:} While effective, especially for long temporal horizons, the proposed methodology also admits some limitations. 
Since NCA relies on local interactions, the information available to each cell is limited by the size of its perception field.
NCA also uses repeated application of the same local update rule, which can cause small prediction errors to accumulate over long rollouts, particularly when the learned update is not sufficiently stable. 
In addition, the performance of NCA can depend on choices such as the perception kernel, update rule, and training rollout length, and these choices may need to change for different types of PDEs. 
Another limitation is that standard NCA does not explicitly include physical constraints such as conservation laws, boundary conditions, or known symmetries unless these are incorporated into the model or training procedure. Therefore, good prediction accuracy on the PDEs considered here does not necessarily imply that the learned dynamics satisfy the underlying physical laws. Finally, because NCA learns the dynamics from data rather than directly using the known PDE, its ability to generalize to parameter regimes, resolutions, or physical settings that are substantially different from those seen during training remains an open question.

Several directions can be explored as a future work. 
Incorporating known physical properties, such as conservation and symmetry, into the NCA update rule could help improve its reliability and generalization. Another important direction is to investigate adaptive or multi-scale perception mechanisms that can capture both local and longer-range interactions without simply increasing the kernel size. Finally, the interpretability results suggest that NCA could be useful not only for predicting PDE solutions but also for identifying the equations that govern them. By examining and sparsifying the learned filters, it may be possible to identify the important differential operators and recover an unknown PDE directly from data.

\paragraph{Use of AI:} The authors would like to acknowledge the use of AI tools such as ChatGPT and Claude AI for improving manuscript quality and building graphical abstract.

\bibliographystyle{cas-model2-names}

\bibliography{cas-refs}

\end{document}